\documentclass{article}

\usepackage[preprint]{neurips_2024}

\usepackage[utf8]{inputenc} 
\usepackage[T1]{fontenc}    
\usepackage{hyperref}       
\usepackage{url}            
\usepackage{booktabs}       
\usepackage{amsfonts}       
\usepackage{nicefrac}       
\usepackage{microtype}      
\usepackage[dvipsnames]{xcolor}
\usepackage{amsmath}

\usepackage{tikz}
\usetikzlibrary{matrix}
\usetikzlibrary{calc}
\usetikzlibrary{shapes, arrows, decorations.pathreplacing, positioning}

\usepackage{multirow}

\usepackage{subcaption}
\usetikzlibrary{arrows.meta, positioning, calc}
\usepackage{amssymb}

\title{SVD-Surgeon: Optimal Singular-Value Surgery for Large Language Model Compression}

\author{%
  Mahmoud Safari\textsuperscript{1} \quad Frank Hutter\textsuperscript{2,3,1} 
  \\[0.4em]
  \textsuperscript{1}University of Freiburg \quad
  \textsuperscript{2}Prior Labs \quad 
  \textsuperscript{3}ELLIS Institute Tübingen
  \\[0.2em]
  \texttt{\{safarim,fh\}@cs.uni-freiburg.de}
}

\begin{document}

\maketitle

\begin{abstract}
Large language models (LLMs) achieve remarkable performance across a wide range of tasks, but their deployment is constrained by substantial memory and compute requirements. Low-rank compression via singular value decomposition (SVD) is an effective remedy, but existing methods focus on how to factorize and which components to keep. 
We introduce SVD-Surgeon, a training-free method that brings the Optimal Brain Surgeon (OBS) framework to the singular-value basis. Treating each singular value as a parameter, it computes a closed-form update of the retained singular values that compensates, to second order in the model loss, for those removed by truncation. The same analysis yields a saliency for choosing which values to prune. As it operates directly on the singular-value factorization, SVD-Surgeon can be layered on top of existing SVD compressors. Applied to SVD-LLM, a leading SVD-based method, it improves the perplexity–compression trade-off on the OPT family and LLaMA 2-7B without any retraining.
\end{abstract}

\section{Introduction}
\label{sec:intro}

Large language models (LLMs) have demonstrated remarkable capabilities across a wide
range of natural language understanding and generation tasks. However, their
deployment remains challenging due to substantial computational and memory
requirements, with state-of-the-art models comprising billions of parameters that
demand significant GPU resources both at inference and fine-tuning time. Reducing
these costs without sacrificing quality has become a central problem for the
practical use of LLMs.

Model compression has emerged as a principled approach to address these demands,
with pruning being among the most widely studied techniques. Structured pruning
removes entire neurons, attention heads, or layers, yielding hardware-friendly
sparse models but at the cost of coarse-grained approximations that often degrade
performance significantly. Unstructured pruning operates at the individual weight
level, achieving fine-grained sparsity but producing irregular patterns that are
difficult to accelerate on modern hardware without specialized kernels.
Semi-structured pruning (e.g.\ $2{:}4$ sparsity) offers a compromise but remains
constrained by fixed patterns that limit flexibility.

Compression using low-rank decomposition such as SVD offers a qualitatively
different approach: by decomposing each weight matrix as $\theta = U\Sigma V^\top$
and truncating to a target rank, one obtains a low-rank approximation that is
naturally deployable via two successive matrix multiplications, without requiring
specialized hardware support. Naive truncation of the smallest singular values is rarely optimal, and a growing
body of work improves on it through importance reweighting and activation whitening
that better align the decomposition with the model's
loss landscape
(Section~\ref{sec:related}).

We introduce \textbf{SVD-Surgeon}, which applies the Optimal Brain Surgeon (OBS)
framework~\citep{hassibi1992second} directly in the basis of singular values.
While prior SVD methods focus on how the low-rank approximation is formed and which components to discard, SVD-Surgeon also
repairs those that remain: given the decomposition produced by any SVD-based
compressor, it excises singular values and adjusts the retained ones to absorb the
induced loss. Concretely, treating each retained singular value as a parameter, it
builds a second-order (Fisher) model of the loss in singular-value coordinates and
derives a closed-form update of the kept singular values that compensates for those
removed by truncation, recovering capacity in a single shot, with no gradient-based
optimization and no post-training fine-tuning. The same analysis yields an OBS saliency for each singular value (i.e. the loss incurred by removing it, accounting for the optimal correction of the others), which provides a principled, loss-aware alternative to magnitude for selecting which components to prune.

\begin{figure}[t]
\centering
\includegraphics[width=\textwidth]{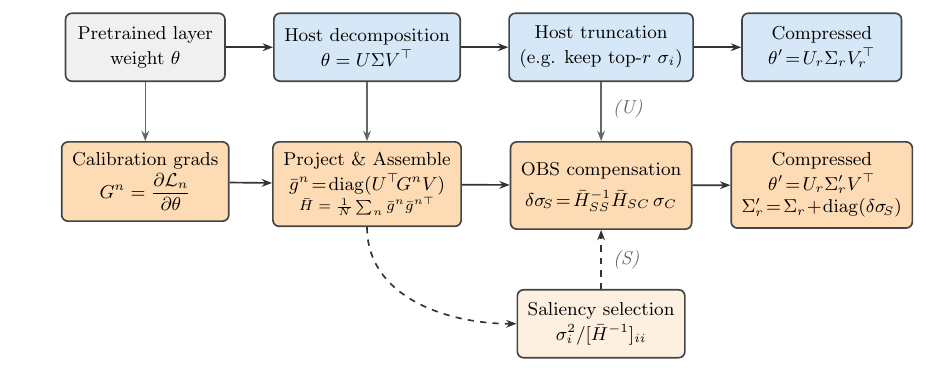}
\caption{Overview of SVD-Surgeon applied to a host SVD compressor
(e.g.\ SVD-LLM). \textbf{Top} (blue): the host pipeline decomposes and
truncates the weight matrix. \textbf{Bottom} (orange): SVD-Surgeon uses
calibration gradients projected into the host's singular-value basis to
assemble a compact Hessian $\bar H$ and compute a closed-form correction
of the retained singular values. Vertical arrows show the information
each SVD-Surgeon stage receives from the host. The update-only variant
(U, solid) inherits the host's pruned set, while the select-and-update
variant (S, dashed) replaces it with a saliency-based selection.}
\label{fig:method}
\end{figure}

Because the derivation assumes only a factorization $\theta=U\Sigma V^\top$ and never
uses orthonormality of $U,V$, SVD-Surgeon can be applied as a corrective layer on top
of a wide range of SVD-based compressors (Figure~\ref{fig:method}). We demonstrate this on SVD-LLM \cite{wangsvd}, a leading SVD-based method, where it improves perplexity across models and compression ratios, with the largest gains under aggressive compression. Post-training fine-tuning is orthogonal and can be layered on top of
SVD-Surgeon, just as it can on any other compression method.

\paragraph{Contributions.}
\begin{itemize}
  \item We bring the Optimal Brain Surgeon framework to the singular-value basis,
  deriving a closed-form, training-free update of the retained singular values that
  compensates, to second order in the loss, for those removed by truncation. The
  same analysis yields a saliency that can be used to select which values to prune.
  \item Since the derivation makes no orthonormality assumption, SVD-Surgeon applies
  on top of a broad class of SVD-based methods. Layered on SVD-LLM, a leading
  SVD-based method, it reduces perplexity across models and compression ratios, with
  the largest gains under aggressive compression.
\end{itemize}

\section{Related Work}
\label{sec:related}

\paragraph{Second-order weight pruning.} Using curvature to guide pruning dates
back to Optimal Brain Damage~\citep{NIPS1989_6c9882bb}, which scores weights by a diagonal
Hessian approximation, and Optimal Brain Surgeon (OBS)~\citep{hassibi1992second}, which uses the
full inverse Hessian to derive both a saliency and a closed-form update of the
surviving weights. Scaling this framework to modern networks motivated a line of
layer-wise approximations: the Optimal BERT Surgeon~\citep{kurtic2022optimal} extended it to
transformers, while Optimal Brain Compression~\citep{frantar2022optimal}, GPTQ~\citep{frantaroptq}, and
SparseGPT~\citep{frantar2023sparsegpt} apply OBS-style closed-form solutions in weight space
for post-training quantization and unstructured pruning of LLMs, using the input
Gram matrix as the layer-wise Hessian. LLM Surgeon~\citep{van2024llm} instead uses a
Kronecker-factored approximate curvature (K-FAC) \citep{martens2015optimizing} for structured and unstructured
weight pruning. All of these operate on the network weights (or their quantized
values). In contrast, SVD-Surgeon applies the OBS framework in the basis of
\emph{singular values}, i.e. the parameters are the $\ell$ singular values of a layer weight matrix and
the relevant Hessian is a small $\ell\times\ell$ matrix obtained by projecting the
gradients onto the singular value space. 


\paragraph{SVD-based LLM compression.} Truncated SVD is a hardware-friendly
alternative to sparsity, and several methods improve on the naive truncation of the
smallest singular values by making it data- or importance-aware.
FWSVD~\citep{hsu2022language} reweights the weight matrix by Fisher importance
before decomposing. DRONE~\citep{chen2021drone} minimizes the output rather than the
weight approximation error using the input distribution, and ASVD~\citep{yuan2023asvd}
scales the weight by activation statistics. SVD-LLM~\citep{wangsvd}, a leading
SVD-based method, introduces a truncation-aware data whitening transform (a Cholesky
factor of the input activation Gram matrix) that aligns singular-value magnitude with
the reconstruction loss, and recovers accuracy through a LoRA-style fine-tuning of the
decomposed factors. OBD-LLM~\citep{li2026optimal} uses a K-FAC factorization of the
task-loss Hessian as a metric, inducing a bidirectional (input- and output-aware)
whitening under which truncated SVD yields a loss-optimal decomposition. All of these
are concerned with how the decomposition is \emph{produced}. SVD-Surgeon instead
performs surgery in the OBS sense on a given decomposition, excising singular
values and repairing the retained ones in closed form. Requiring no orthonormality of the factors, it composes on top of
many such methods, which we demonstrate on SVD-LLM.

\section{SVD-Surgeon}

\subsection{Background: Optimal Brain Surgeon}
Consider a pretrained model whose parameters $\theta$ have converged near a
minimum of the loss $\mathcal{L}$. Since the gradient is then negligible, the
leading change in the loss under a perturbation $\delta\theta$ is second order,
\begin{equation}
        \delta \mathcal{L} \approx  \frac{1}{2} \delta\theta H \delta\theta \label{eq:deltal}
\end{equation}
where $\delta\theta\,H\,\delta\theta$ denotes the quadratic form built from the
Hessian $H$ of $\mathcal{L}$. 
Given a partition of the parameters into a retained set
$\theta_S$ and a complementary set $\theta_C$ to be pruned, 
OBS uses \eqref{eq:deltal} to answer the following
question: if we prune (set to zero) the components $\theta_C$, how do we shift the surviving entries $\delta\theta_S$ to optimally compensate for the induced increase in the loss?


In the standard treatment $\theta$ is vectorized, so $H$ is an ordinary square
matrix and~\eqref{eq:deltal} a familiar quadratic form. We
instead keep $\theta$ in its native matrix form, which makes the reduction
to the SVD basis in the next subsection transparent. With $\theta$ a weight matrix, $H$ is the
fourth-order tensor
\begin{equation}
  H_{ij,kl} = \frac{\partial^{2}\mathcal{L}}{\partial\theta_{ij}\,\partial\theta_{kl}},
  \qquad
  \delta\theta H\delta\theta = \sum_{i,j,k,l}\delta\theta_{ij}\,H_{ij,kl}\,\delta\theta_{kl}.
  \label{eq:hessian}
\end{equation}

\subsection{Compression in the SVD basis}
\label{sec:svd-basis}

Classical OBS operates on the components, whether individual or entire rows or columns, of the matrix $\theta$. We instead apply the same framework to {\it singular values} of $\theta$.
Write a general decomposition of the weight matrix $\theta\in\mathbb{R}^{m\times n}$ as
\begin{equation}
        \theta = \sum_{i=1}^\ell \sigma_i u_i v_i^\top = U \Sigma V^\top \label{theta}
\end{equation}
where $\ell=\min(m,n)$, $\Sigma=\mathrm{diag}(\sigma_1,\dots,\sigma_\ell)$, and
$u_i,v_i$ are the columns of $U\in\mathbb{R}^{m\times\ell}$ and
$V\in\mathbb{R}^{n\times\ell}$. We do not require $U$ and $V$ to be orthonormal, so
the standard SVD ($U^\top U=V^\top V=I$) is a special case. For simplicity, and with a slight abuse of terminology, we continue to call the $\sigma_i$ singular values and the $u_, v_i$ singular directions even in this general, non-orthonormal case.

In this basis compression amounts to discarding singular triplets, i.e. setting $\sigma_i=0$
removes the rank-one term $\sigma_i u_i v_i^\top$. We 
partition the index set
accordingly into the $r$ retained values $\sigma_S$ and the $\ell-r$ pruned
values $\sigma_C$, with $r$ the target rank. Pruning imposes
$\delta\sigma_C=-\sigma_C$, and the OBS question becomes: what is the optimal
update $\delta\sigma_S$ of the retained singular values? In principle the
directions $u_i,v_i$ could also be relaxed, but here we update only the singular
values, which is inexpensive and, as we show, already effective.

\subsection{Reduction to singular-value space}
\label{sec:reduction}
We restrict attention to changes in the singular values alone, keeping the
singular directions $U,V$ fixed. This means the weight variation takes the form
$\delta\theta = U\,\delta\Sigma\, V^\top$ with
$\delta\Sigma = \operatorname{diag}(\delta\sigma)$. Substituting this into the
quadratic model~\eqref{eq:deltal} and expanding in indices,
\begin{align}
2\,\delta\mathcal{L} = \sum_{i,j,k,l} \delta\theta_{ij}\,H_{ij,kl}\,\delta\theta_{kl} 
&= \sum_{i,j,k,l} (U\delta\Sigma V^\top)_{ij}\, H_{ij,kl}\, (U\delta\Sigma V^\top)_{kl} \\
&= \sum_{\substack{i,j,k,l\\ p,q}}
   U_{ip}\,\delta\Sigma_{pp}\,V^\top_{pj}\, H_{ij,kl}\,
   U_{kq}\,\delta\Sigma_{qq}\,V^\top_{ql} \\
&= \sum_{\substack{i,j,k,l\\ p,q}}
   \delta\sigma_{p}\, U^\top_{pi} V_{jp}\, H_{ij,kl}\, U^\top_{qk} V_{lq}\, \delta\sigma_{q} \\
&= \,\sum_{p,q} \delta\sigma_{p}\, \bar H_{pq}\, \delta\sigma_{q}
   \;=\; \delta\sigma^\top \bar H\, \delta\sigma ,
\end{align}
where in the third line we used that $\delta\Sigma$ is diagonal, writing
$\delta\Sigma_{pp}=\delta\sigma_p$, and
collected the scalar factors $\delta\sigma_p$, $\delta\sigma_q$ to the
outside. Also, in the last line we defined
\begin{equation}
\bar H_{pq} \;\equiv\; \sum_{i,j,k,l} U^\top_{pi}\, V_{jp}\, H_{ij,kl}\, U^\top_{qk}\, V_{lq}.
\label{eq:hbar}
\end{equation}
The fourth-order form in $\delta\theta$ then collapses to an ordinary
quadratic form in the $\ell$-vector $\delta\sigma$:
\begin{equation}
  \delta\mathcal{L}
  \;\approx\;
  \frac{1}{2}\,\delta\sigma^\top \bar H\,\delta\sigma.
  \label{eq:deltal_sigma}
\end{equation}
Here $\sigma$ stacks all $\ell$ singular values, comprising the retained
$\sigma_S$ and the pruned $\sigma_C$ introduced above. The
reduction~\eqref{eq:deltal_sigma} is the key practical simplification:
$\bar H$ is only $\ell\times\ell$ (at most $\min(m,n)$ in each dimension),
whereas the weight-space Hessian in~\eqref{eq:hessian} is $mn\times mn$.



\subsection{Fisher approximation of the Hessian}
\label{sub:fisher}
We assume block-diagonality of the Hessian across layers, so that each
layer's projected Hessian $\bar H$ can be estimated independently. Near
convergence the per-layer Hessian is well approximated by the empirical
Fisher information, a sum of outer products of per-sample gradients
$G^{n}=\partial\mathcal{L}_n/\partial\theta$,
\begin{equation}
  H_{ij,kl} \;\approx\; \frac{1}{N}\sum_{n=1}^{N} G^{n}_{ij}\,G^{n}_{kl},
  \label{eq:fisher}
\end{equation}
where $N$ is the number of calibration samples. Inserting~\eqref{eq:fisher}
into the definition of $\bar H$~\eqref{eq:hbar} and carrying out the
contractions over $i,j,k,l$:
\begin{align}
\bar H_{pq}
&= \frac{1}{N}\sum_n \sum_{i,j,k,l}
   U_{pi}^\top\,V_{jp}\,G^n_{ij}\,G^n_{kl}\,U_{qk}^\top\,V_{lq} \notag\\
&= \frac{1}{N}\sum_n
   \bigl(U^\top G^n V\bigr)_{pp}\;
   \bigl(U^\top G^n V\bigr)_{qq} \notag\\
&= \frac{1}{N}\sum_n \bar g^n_p\;\bar g^n_q,
\label{eq:Hbar_expand}
\end{align}
where in the second line, since $p$ and $q$ each appear in only one
factor, the sum over all matrix indices collapses to the product of two
diagonal entries of $U^\top G^n V$, leading to a further simplification
of the projected Hessian. Defining the projected
gradient $\bar g^n_p \equiv (U^\top G^n V)_{pp}$, this reads in matrix form
\begin{equation}
  \bar H \;=\; \frac{1}{N}\sum_{n=1}^{N}\bar g^{n}\,\bar g^{n\top},
  \qquad
  \bar g^n \;\equiv\; \operatorname{diag}\bigl(U^\top G^{n} V\bigr).
  \label{eq:Hbar_fisher}
\end{equation}
In fact, by the chain rule and
$\partial\theta_{ij}/\partial\sigma_p = U_{ip}\,V_{jp}$, each component
$\bar g^{n}_p$ is simply the singular-value gradient
$\partial\mathcal{L}_n/\partial\sigma_p$. These derivatives are not standard
operations in automatic-differentiation frameworks. However, such frameworks
readily provide the full gradient matrices $G^n$, from which $\bar g^n$ is
obtained via a single matrix product $U^\top G^n V$ and diagonal extraction.
Similarly, $\bar H$ coincides with the exact loss Hessian in singular-value
coordinates, $\bar H_{pq} =
\partial^2\mathcal{L}/\partial\sigma_p\,\partial\sigma_q$, but the Fisher
outer-product form~\eqref{eq:Hbar_fisher} avoids computing these second
derivatives altogether, assembling the $\ell\times\ell$ Hessian entirely from
first-order information.

\subsection{Optimal singular-value update}
\label{sec:update}

In order to obtain the optimal update $\delta\sigma_S$ of the retained singular values after pruning, we
partition $\bar H$ conformally with $(\sigma_S,\sigma_C)$ into blocks
$\bar H_{SS}$, $\bar H_{CC}$ and $\bar H_{SC}=\bar H_{CS}^\top$, so that
\begin{equation}
  \delta\mathcal{L}
  \;=\; \tfrac{1}{2}\,\delta\sigma_S^\top \bar H_{SS}\,\delta\sigma_S
  \;+\; \tfrac{1}{2}\,\delta\sigma_C^\top \bar H_{CC}\,\delta\sigma_C
  \;+\; \delta\sigma_S^\top \bar H_{SC}\,\delta\sigma_C.
  \label{eq:deltal_sigma_split}
\end{equation}
Pruning fixes $\delta\sigma_C=-\sigma_C$. Rather than enforcing this with a
Lagrange multiplier, we equivalently substitute it directly and minimize the resulting
unconstrained objective over $\delta\sigma_S$,
\begin{equation}
  \delta\sigma_S^{\star}
  \;=\; \underset{\delta\sigma_S}{\mathrm{argmin}}
  \Big(\tfrac{1}{2}\,\delta\sigma_S^\top  \bar H_{SS}\,\delta\sigma_S
  \;-\; \delta\sigma_S^\top \bar H_{SC}\,\sigma_C\Big).
\end{equation}
Notice that we have dropped the term $\tfrac{1}{2}\,\sigma_C^\top  \bar H_{CC}\,\sigma_C$ which is independent of $\delta\sigma_S$. Setting the derivative with respect to $\delta\sigma_S$ to zero gives the optimal
compensation in a simple closed form,
\begin{equation}
\delta\sigma_S^{\star} \;=\; \bar H_{SS}^{-1}\,\bar H_{SC}\,\sigma_C
\label{eq:delta_sigma}
\end{equation}
which prescribes how the surviving singular values should shift to absorb, to
second order, the effect of removing $\sigma_C$.

\subsection{Saliency scores and pruning selection}
\label{sec:saliency}
Substituting the optimal update~\eqref{eq:delta_sigma} back
into~\eqref{eq:deltal_sigma_split} yields the loss increase induced by
removing $\sigma_C$ and optimally compensating $\sigma_S$. Expanding and
simplifying, we get:
\begin{align}
\delta\mathcal{L}
&= \tfrac{1}{2}\,\delta\sigma_S^\top \bar H_{SS}\,\delta\sigma_S
   - \sigma_C^\top \bar H_{CS}\,\delta\sigma_S
   + \tfrac{1}{2}\,\sigma_C^\top \bar H_{CC}\,\sigma_C \\[4pt]
&= \tfrac{1}{2}\,\sigma_C^\top \bar H_{CS}\,\bar H_{SS}^{-1}
   \bar H_{SS}\,\bar H_{SS}^{-1}\bar H_{SC}\,\sigma_C
   - \sigma_C^\top \bar H_{CS}\,\bar H_{SS}^{-1}\bar H_{SC}\,\sigma_C
   + \tfrac{1}{2}\,\sigma_C^\top \bar H_{CC}\,\sigma_C \\[4pt]
&= \tfrac{1}{2}\,\sigma_C^\top
   \bigl(\bar H_{CC} - \bar H_{CS}\,\bar H_{SS}^{-1}\bar H_{SC}\bigr)\,\sigma_C.
\end{align}
The matrix in parentheses is the Schur complement of $\bar H_{SS}$ in
$\bar H$. The first term $\bar H_{CC}$ captures the naive cost of
removing $\sigma_C$, while the second term accounts for the reduction
afforded by optimally adjusting $\sigma_S$. By the standard block-inversion identity,
$[\bar H^{-1}]_{CC} =
(\bar H_{CC} - \bar H_{CS}\,\bar H_{SS}^{-1}\bar H_{SC})^{-1}$, so the loss
increase can equivalently be written as
\begin{equation}
\delta\mathcal{L}
\;=\; \tfrac{1}{2}\,\sigma_C^\top
      \bigl([\bar H^{-1}]_{CC}\bigr)^{-1}\sigma_C,
\label{eq:saliency_schur}
\end{equation}
which recovers, in equivalent form, the OBS result obtained via Lagrange
multipliers~\citep{kurtic2022optimal}. For the special case of removing a
single singular value $\sigma_i$, this collapses to the familiar OBS
saliency,
\begin{equation}
  \delta\mathcal{L} \;=\; \frac{\sigma_i^{2}}{2\,[\bar H^{-1}]_{ii}},
  \qquad i\in C,
  \label{eq:saliency}
\end{equation}
which scores each triplet by the loss it would incur if pruned, accounting
for the optimal correction of the survivors.

This leads to two variants of our method, both of which apply the same
compensation~\eqref{eq:delta_sigma} but differ only in how the pruned set $C$
is chosen:
\begin{itemize}
  \item \textbf{Update-only (U).} $C$ is inherited from the host's
        criterion (e.g.\ the smallest $\sigma_i$), and we apply the compensation on top of it.
  \item \textbf{Select-and-update (S).} We rank triplets by the
        saliency~\eqref{eq:saliency}, take the lowest-scoring as $C$, and
        then apply the compensation.
\end{itemize}
When $\bar H\approx I$, the saliency reduces to $\sigma_i^2/2$, recovering
magnitude-based selection. 

\subsection{Application to SVD-LLM}
SVD-Surgeon assumes only a factorization $\theta=U\Sigma V^\top$ and does not use
orthonormality of $U,V$, so it can be applied on top of any method that compresses
by truncating such a factorization. We instantiate it on SVD-LLM, a natural host:
conceptually simple yet among the strongest SVD-based compressors, and, since its
whitening makes one of the factors non-orthonormal, a non-trivial test of this generality.

SVD-LLM is built around truncation-aware data whitening. It takes the layer
reconstruction error as the compression loss,
\begin{equation}
    \min_{\theta'}\;\|\theta X - \theta' X\|_F,
\end{equation}
where $X$ stacks the calibration activations, and whitens the inputs using a
Cholesky factor $S$ of the activation Gram matrix, $XX^\top = SS^\top$. Since
$(S^{-1}X)(S^{-1}X)^\top = I$, the loss equals
$\|(\theta S-\theta'S)S^{-1}X\|_F=\|\theta S-\theta'S\|_F$, a plain Frobenius
distance between the whitened weights, whose optimal rank-$r$ solution is the
truncated SVD of $\theta S$. Concretely, one computes $\theta S = U\Sigma\tilde
V^\top$, truncates $\Sigma$ to its top $r$ values, and maps back through $S^{-1}$,
\begin{equation}
    \theta' = U\,\mathrm{Trunc}(\Sigma)\,\tilde V^\top S^{-1}.
\end{equation}
Cast in the form $\theta=U\Sigma V^\top$ of Eq.~\eqref{theta}, the left factor $U$
remains orthonormal while the right factor $V^\top \equiv \tilde V^\top S^{-1}$
absorbs the inverse whitening transform and is no longer orthonormal. We apply
SVD-Surgeon's compensation~\eqref{eq:delta_sigma} to this decomposition, inheriting
SVD-LLM's truncation set in the update-only variant or re-selecting it
via~\eqref{eq:saliency} in select-and-update.

\section{Experiments}
\label{sec:experiments}

\subsection{Setup}
\label{sub:setup}

\paragraph{Models and data.} We evaluate on the OPT family (1.3B, 2.7B, 6.7B) and LLaMA-2-7B,
reporting perplexity ($\downarrow$) on WikiText-2 for all four models
and on C4 for OPT-1.3B and OPT-2.7B. We sweep the
compression ratio $\rho$ (the fraction of parameters removed) from $20\%$ to $80\%$, with
emphasis on the high-compression regime, where truncation is most damaging. For an $m\times n$ weight, a rank-$r$ factorization stores $r(m+n)$ parameters, so $\rho = 1 - r(m+n)/mn$. A
target ratio $\rho$ thus corresponds to rank $r = (1-\rho)\,mn/(m+n)$, rounded to the nearest
integer per layer.

\paragraph{Calibration.} SVD-Surgeon is single-shot: it estimates the Fisher
information $\bar H$ from a calibration set and produces the compressed model in one
pass, with no gradient-based optimization. We draw this set from the same source as
the host's whitening data but keep it separate, since accurate Hessian estimation
requires substantially more samples ($N$) than whitening ($N_{\text{cal}}$).
Forming $\bar H$ is the most expensive step of the pipeline, but it is a one-time,
offline computation that parallelizes across batches.

\paragraph{Implementation.} Before the inversions in~\eqref{eq:delta_sigma}
and~\eqref{eq:saliency} we add a diagonal damping, $d_S$ and $d$, to $\bar H_{SS}$ and $\bar H$ respectively, and we scale the compensation
update~\eqref{eq:delta_sigma} by a factor $\lambda$ to account for the approximate
Hessian. For efficiency, we retain only the leading $r+\alpha(\ell-r)$ block of
$\bar H$, corresponding to the top $r$ singular values plus a fraction $\alpha$ of the
remainder, and discard the rest.
The hyperparameter settings for $\lambda$, $N$, the two damping coefficients $d_S,d$, and
$\alpha$ are reported in Appendix~\ref{app:hparams}. All experiments were conducted on a single NVIDIA H200 GPU. \footnote{Code:
\url{https://github.com/mahmoud-safari/SVD-Surgeon}}

\paragraph{Baseline.} We layer SVD-Surgeon on SVD-LLM and compare against SVD-LLM with
truncation-aware whitening only: its LoRA-style fine-tuning recovery is orthogonal
and can be applied on top of any method, including SVD-Surgeon, just as SVD-LLM
applies it to its own decomposition. We report both variants of our method:
\emph{update-only} (U), which inherits the host's truncation set and applies the
compensation~\eqref{eq:delta_sigma}, and \emph{select-and-update} (S), which
additionally re-selects the pruned set via the saliency~\eqref{eq:saliency}.

\subsection{Results}

\paragraph{Perplexity.} Table~\ref{tab:ppl_vs_ratio_wiki} reports WikiText-2 perplexity for SVD-Surgeon layered on
SVD-LLM (whitening only), across compression ratios for OPT-1.3B, OPT-2.7B,
OPT-6.7B, and LLaMA-2-7B.  the corresponding
perplexity curves are plotted in Appendix~\ref{app:wikitext2}. C4 results (OPT-1.3B, OPT-2.7B) are given in
Table~\ref{tab:ppl_vs_ratio_c4}. We report both variants, update-only (U) and
select-and-update (S). Across all four models, SVD-Surgeon improves on
SVD-LLM, and the gains grow with the compression ratio. under mild compression there is little to compensate, while
under aggressive compression SVD-Surgeon prevents the steep degradation that SVD-LLM
suffers (e.g. WikiText-2 perplexity on OPT-6.7B, 944.57 $\to$ 46.36 at ratio $0.7$). Most
of this improvement comes from the closed-form update. Re-selecting the pruned set
via the saliency (S) adds a smaller, further gain in most settings.

SVD-LLM is deterministic given fixed calibration data. The Fisher estimate
in SVD-Surgeon introduces small variance through CUDA non-determinism in
the gradient computation; for OPT models we report the mean over 3 seeds
(standard deviations are given in Appendix~\ref{app:wikitext2}), while for
LLaMA-2-7B the results were identical across seeds and we report a single
value.
\begin{table}[t]
\centering
\small
\caption{WikiText-2 perplexity ($\downarrow$) vs.\ compression ratio across models. Dense perplexity: OPT-1.3B\,=\,14.62, OPT-2.7B\,=\,12.47,
OPT-6.7B\,=\,10.86, LLaMA-2-7B\,=\,5.47. 
\textbf{Bold}: best; \underline{underline}: second best.}
\label{tab:ppl_vs_ratio_wiki}
\begin{tabular}{llccccccc}
\toprule
Model & Method & 0.2 & 0.3 & 0.4 & 0.5 & 0.6 & 0.7 & 0.8 \\
\midrule
\multirow{3}{*}{OPT-1.3B}
  & SVD-LLM        & 17.82 & 20.69 & 27.28 & 47.79 & 140.82 & 654.89 & 4206.70 \\
  & SVD-Surgeon (U)    & \underline{17.50} & \underline{19.62} & \underline{23.37} & \textbf{31.29} & \underline{53.63} & \underline{168.48} & \underline{1728.88} \\
  & SVD-Surgeon (S)    & \textbf{17.49} & \textbf{19.59} & \textbf{23.31} & \underline{31.39} & \textbf{53.46} & \textbf{163.41} & \textbf{1587.61} \\
\midrule
\multirow{3}{*}{OPT-2.7B}
  & SVD-LLM        & 15.21 & 17.80 & 23.39 & 40.19 & 125.86 & 877.25 & 4584.67 \\
  & SVD-Surgeon (U)    & \underline{14.93} & \underline{16.97} & \underline{20.91} & \underline{29.15} & \underline{51.43} & \underline{149.52} & \underline{1041.95} \\
  & SVD-Surgeon (S)    & \textbf{14.83} & \textbf{16.87} & \textbf{20.85} & \textbf{29.11} & \textbf{51.33} & \textbf{146.67} & \textbf{984.74} \\
\midrule
\multirow{3}{*}{OPT-6.7B}
  & SVD-LLM        & 12.05 & 13.08 & 15.27 & 21.22 & 53.23 & 944.57 & 6777.57 \\
  & SVD-Surgeon (U)    & \textbf{12.00} & \underline{12.81} & \underline{14.25} & \underline{17.02} & \underline{23.76} & \underline{47.27} & \underline{316.47} \\
  & SVD-Surgeon (S)    & \underline{12.01} & \textbf{12.80} & \textbf{14.22} & \textbf{16.90} & \textbf{23.39} & \textbf{46.36} & \textbf{279.88} \\
\midrule
\multirow{3}{*}{LLaMA-2-7B}
  & SVD-LLM        & 8.38 & 10.67 & 16.15 & 33.28 & 89.97 & 253.40 & 570.44 \\
  & SVD-Surgeon (U)    & \underline{8.34} & \underline{10.52} & \underline{15.71} & \underline{31.49} & \underline{84.04} & \underline{241.71} & \underline{549.14} \\
  & SVD-Surgeon (S)    & \textbf{8.20} & \textbf{10.36} & \textbf{15.59} & \textbf{31.14} & \textbf{82.50} & \textbf{232.76} & \textbf{531.62} \\
\bottomrule
\end{tabular}
\end{table}
\begin{table}[h]
\centering
\small
\caption{C4 perplexity ($\downarrow$) vs.\ compression ratio. Dense perplexity: OPT-1.3B\,=\,15.68, OPT-2.7B\,=\,14.06. 
\textbf{Bold}: best; \underline{underline}: second best.}
\label{tab:ppl_vs_ratio_c4}
\begin{tabular}{llccccccc}
\toprule
Model & Method & 0.2 & 0.3 & 0.4 & 0.5 & 0.6 & 0.7 & 0.8 \\
\midrule
\multirow{3}{*}{OPT-1.3B}
  & SVD-LLM          & 20.08 & 24.88 & 37.83 & 84.20 & 329.17 & 1360.97 & 4220.29 \\
  & SVD-Surgeon (U)   & \underline{18.99} & \textbf{21.68} & \underline{27.01} & \underline{38.72} & \textbf{74.61} & \underline{293.13} & \underline{3562.08} \\
  & SVD-Surgeon (S)   & \textbf{18.98} & \underline{21.69} & \textbf{26.89} & \textbf{38.54} & \underline{75.41} & \textbf{278.78} & \textbf{2915.40} \\
\midrule
\multirow{3}{*}{OPT-2.7B}
  & SVD-LLM          & 17.60 & 21.48 & 32.02 & 63.00 & 211.21 & 1256.38 & 7113.83 \\
  & SVD-Surgeon (U)   & \underline{16.73} & \underline{18.99} & \underline{23.72} & \underline{34.19} & \textbf{63.18} & \textbf{220.83} & \underline{4155.44} \\
  & SVD-Surgeon (S)   & \textbf{16.70} & \textbf{18.91} & \textbf{23.56} & \textbf{34.00} & \underline{63.28} & \underline{228.09} & \textbf{3664.16} \\
\bottomrule
\end{tabular}
\end{table}

\begin{figure}[t]
\centering
\begin{subfigure}[b]{0.48\textwidth}
    \centering
    \includegraphics[width=\textwidth]{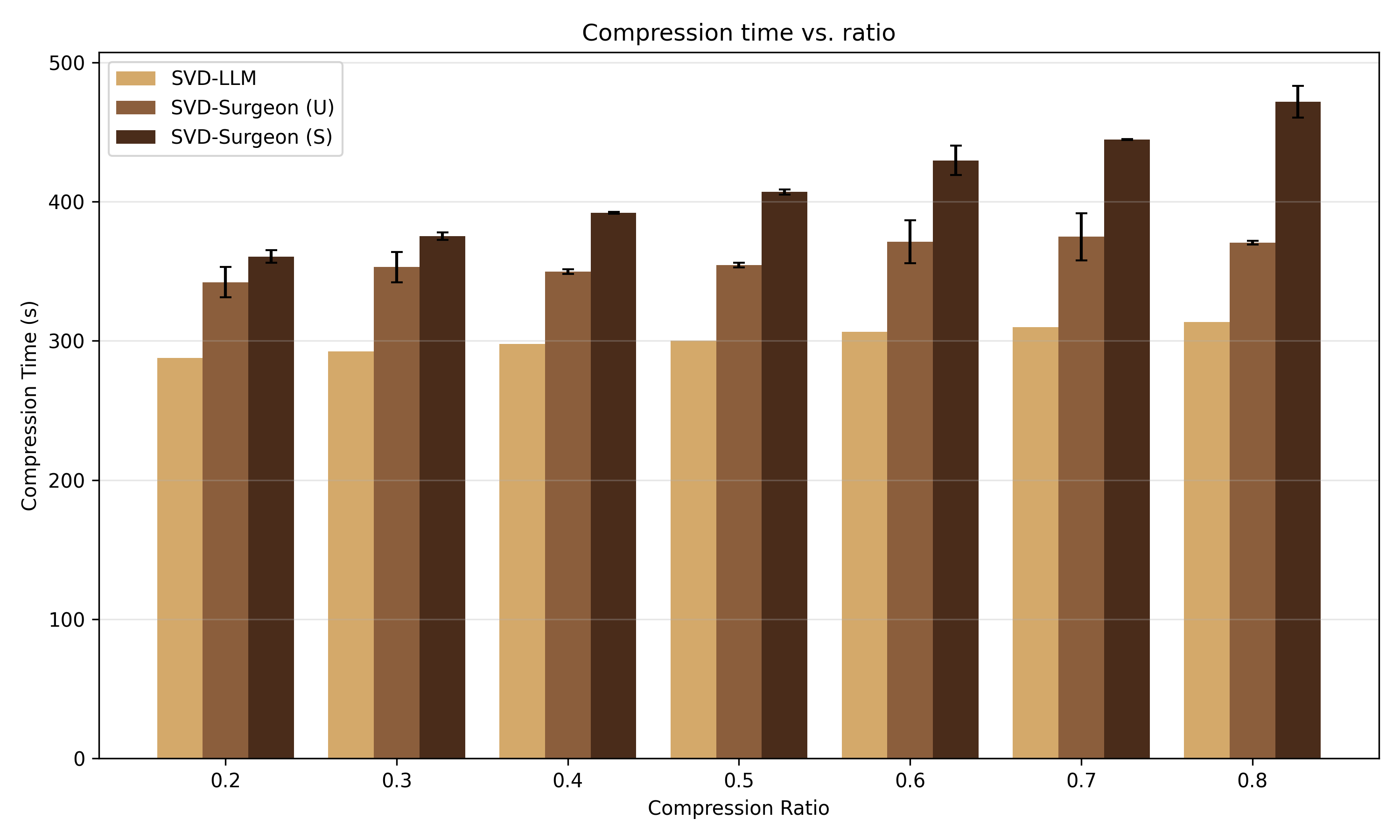}
    \caption{LLaMA-2-7B}
    \label{fig:time_llama}
\end{subfigure}
\hfill
\begin{subfigure}[b]{0.48\textwidth}
    \centering
    \includegraphics[width=\textwidth]{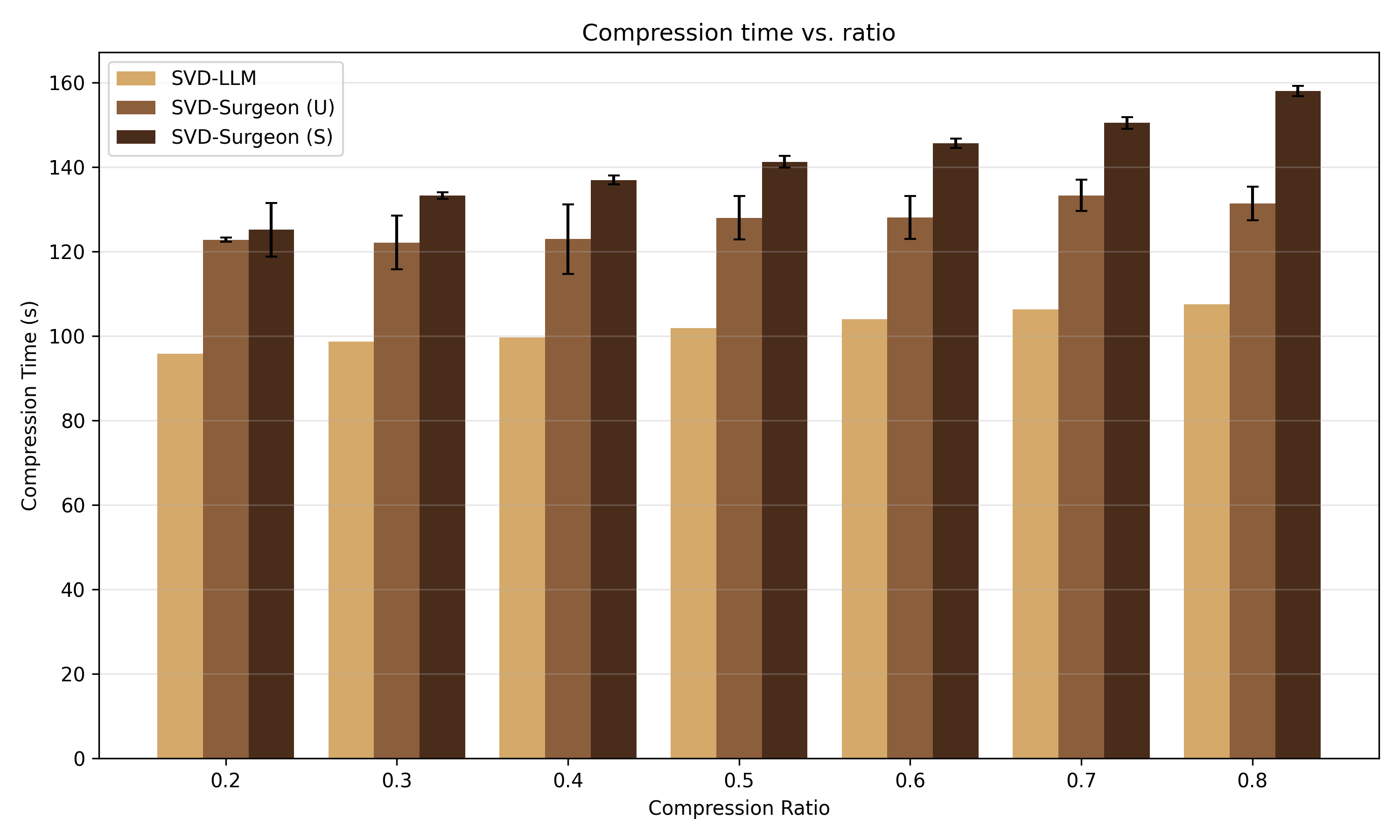}
    \caption{OPT-2.7B}
    \label{fig:time_opt}
\end{subfigure}
\caption{Wall-clock compression time (seconds), measured on a single H200 GPU, across compression ratios.
The pattern is similar for the remaining models.}
\label{fig:compression_time}
\end{figure}

\paragraph{Compression time.}
Figure~\ref{fig:compression_time} shows wall-clock compression time across
ratios for LLaMA-2-7B and OPT-2.7B (the pattern is similar for the remaining
models). This is defined as the time of the pruning algorithm itself,
including loading the precomputed $\bar H$ from disk for SVD-Surgeon, but
excluding model and calibration-data loading, which are shared across all
methods. The update-only variant adds modest overhead as it only solves a single linear system against $\bar H_{SS}$. The select-and-update
variant is more expensive because it requires a pseudoinverse of the full
(block-truncated) $\bar H$ for the saliency scores. Assembling $\bar H$
itself costs roughly $7.6$\,s per calibration sample for LLaMA-2-7B
($2.6$\,s for OPT-2.7B). This is a one-time computation, parallelisable
across samples and layers, whose result is reused across all compression
ratios.

\section{Conclusion}
\label{sec:conclusion}

We introduced SVD-Surgeon, a method that brings the Optimal Brain Surgeon
framework to the singular-value basis. By treating singular values as
first-class parameters and building a compact, Fisher-estimated Hessian in
that coordinate system, SVD-Surgeon derives a closed-form update of the
retained singular values that compensates for those removed by truncation, as
well as a compensation-aware saliency that can replace magnitude-based
selection. The entire procedure is training-free, requires no iterative
optimization, and, because it assumes no orthonormality of the
factors, composes directly on top of existing SVD compressors. Applied to
SVD-LLM, SVD-Surgeon improves the perplexity-compression trade-off across
models and compression ratios, with the largest gains in the aggressive
regime where standard truncation is most damaging.

\paragraph{Limitations.}
The current evaluation measures perplexity on two benchmarks (WikiText-2
and C4); validating on additional
hosts, model families, and downstream tasks would strengthen the generality
claim. By design, the method updates only the singular values while holding
the directions $U,V$ fixed. This is a deliberate simplification. The
$\ell$-dimensional $\sigma$-space is far smaller than the full parameter
space, keeping both the derivation and the computation tractable, and the
experiments show it can have a substantial effect, particularly under
aggressive compression. Whether additionally varying $U$ and $V$ within the
same second-order framework yields further gains that justify the added computational cost is an open question.
Estimating the Fisher information requires a forward and backward pass over a
calibration set for each layer, which can be expensive for large models, although in
practice this is a one-time, offline computation that parallelises across both
layers and samples. Finally, SVD-Surgeon
introduces several hyperparameters (Appendix~\ref{app:hparams}). Default
values proved robust across most configurations, but the interaction between
these settings and factors such as model scale or Fisher accuracy is not yet
fully understood. A principled selection scheme would make the method more
plug-and-play.

\paragraph{Future work.}
Several directions follow naturally. SVD-Surgeon could be layered on other
SVD-family compressors to test composability more
broadly, and combined with post-hoc fine-tuning to measure whether the two
recovery mechanisms stack, or paired
with quantization of the corrected factors for additional compression. The per-component saliency is expressed in units of loss and is therefore
comparable across layers, which could enable global, cross-layer rank
allocation as a natural extension. More ambitiously, relaxing the fixed-direction assumption
to allow joint updates of $U$, $\Sigma$, and $V$ within the same second-order
framework could close the gap left by freezing the singular directions.

\section*{Acknowledgements}

This research was funded by the Deutsche Forschungsgemeinschaft (DFG, German Research Foundation) under grant number 539134284, through EFRE (FEIH\_2698644) and the state of Baden-Württemberg.
\begin{center}
\includegraphics[width=0.3\textwidth]{figures/BaWue_Logo_Standard_rgb_pos.png} ~~~ \includegraphics[width=0.3\textwidth]{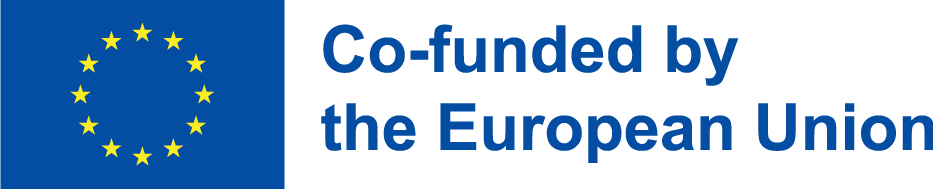} 
\end{center}
Frank Hutter acknowledges the financial support of the Hector Foundation.




\bibliography{main}
\bibliographystyle{plain}


\appendix

\section{Hyperparameter Settings}
\label{app:hparams}

Tables~\ref{tab:hparams_wiki} and \ref{tab:hparams_c4} list, respectively, the hyperparameters introduced in Section~\ref{sub:setup}. The compensation scaling $\lambda$, the damping coefficients $d_S$ and $d$
(expressed as fractions of the mean diagonal of $\bar H_{SS}$ and $\bar H$,
respectively), the block-truncation fraction $\alpha$, and the number of Fisher samples $N$. For each model the same
settings are used across all compression ratios, and only across models do they differ.
The three hyperparameters $\lambda, d_S, d$ were chosen via light manual exploration starting from natural defaults (e.g. $\lambda=1$) rather than aggressive tuning. The final values were held fixed across all compression ratios, which preserves the single-shot character of the method. Notice that the compensation scaling $\lambda$ has been set to $1$ for all OPT models and to $0.1$ for Llama 2-7B.

The block-truncation fraction $\alpha$ controls a trade-off between accuracy
and cost: larger values retain a greater portion of $\bar H$, improving the
fidelity of the compensation update at the expense of higher computational
overhead for Fisher collection and the inversion and matrix operations that follow. We fix
$\alpha=0.3$ across all models and compression ratios, which we found to
offer a good balance. The number of Fisher samples $N$ was chosen large enough for the projected
Hessian $\bar H$ to be well converged
but was not tuned as a hyperparameter. The hyperparameters of SVD-LLM itself, including the
number of whitening calibration samples ($N_{\mathrm{cal}}=256$), are
kept identical across both methods to ensure a fair comparison. 


\begin{table}[h]
\centering
\caption{Hyperparameter settings for SVD-Surgeon on WikiText-2. Settings are fixed across all
compression ratios for a given model.}
\label{tab:hparams_wiki}
\begin{tabular}{lccccc}
\toprule
Model & $\lambda$ & $d_S$ & $d$ & $\alpha$ & $N$ \\
\midrule
OPT-1.3B    & $1$ & $10^{-5}$ & 1 & $0.3$ & 16384 \\
OPT-2.7B    & $1$ & $10^{-5}$ & 1 & $0.3$ & 24576 \\
OPT-6.7B    & $1$ & $10^{-5}$ & $10^{-7}$ & $0.3$ & 32768 \\
LLaMA-2-7B  & $0.1$ & $10^{-5}$ & $0.1$ & $0.3$ & 32768 \\
\bottomrule
\end{tabular}
\end{table}

\begin{table}[h]
\centering
\caption{Hyperparameter settings for SVD-Surgeon on C4. Settings are fixed across all
compression ratios for a given model.}
\label{tab:hparams_c4}
\begin{tabular}{lccccc}
\toprule
Model & $\lambda$ & $d_S$ & $d$ & $\alpha$ & $N$ \\
\midrule
OPT-1.3B    & $1$ & $10^{-2}$ & $1$ & $0.3$ & 16384 \\
OPT-2.7B    & $1$ & $10^{-5}$ & $10^{-5}$ & $0.3$ & 24576 \\
\bottomrule
\end{tabular}
\end{table}

\section{Extended WikiText-2 Results}
\label{app:wikitext2}

\begin{figure}[t]
\centering
\begin{subfigure}[b]{0.48\textwidth}
    \centering
    \includegraphics[width=\textwidth]{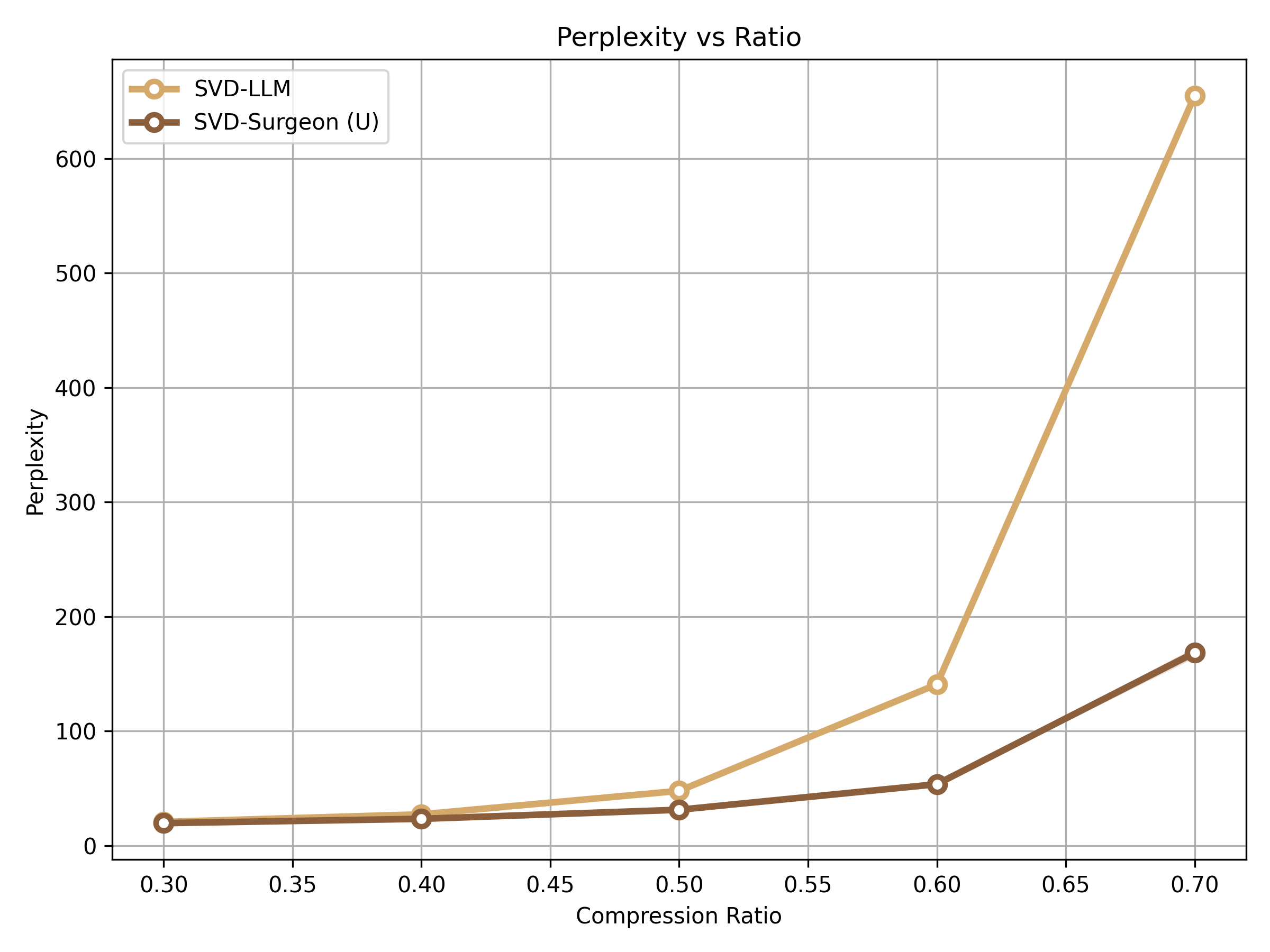}
    \caption{OPT-1.3B}
    \label{fig:ppl_curve_opt_1.3b}
\end{subfigure}
\hfill
\begin{subfigure}[b]{0.48\textwidth}
    \centering
    \includegraphics[width=\textwidth]{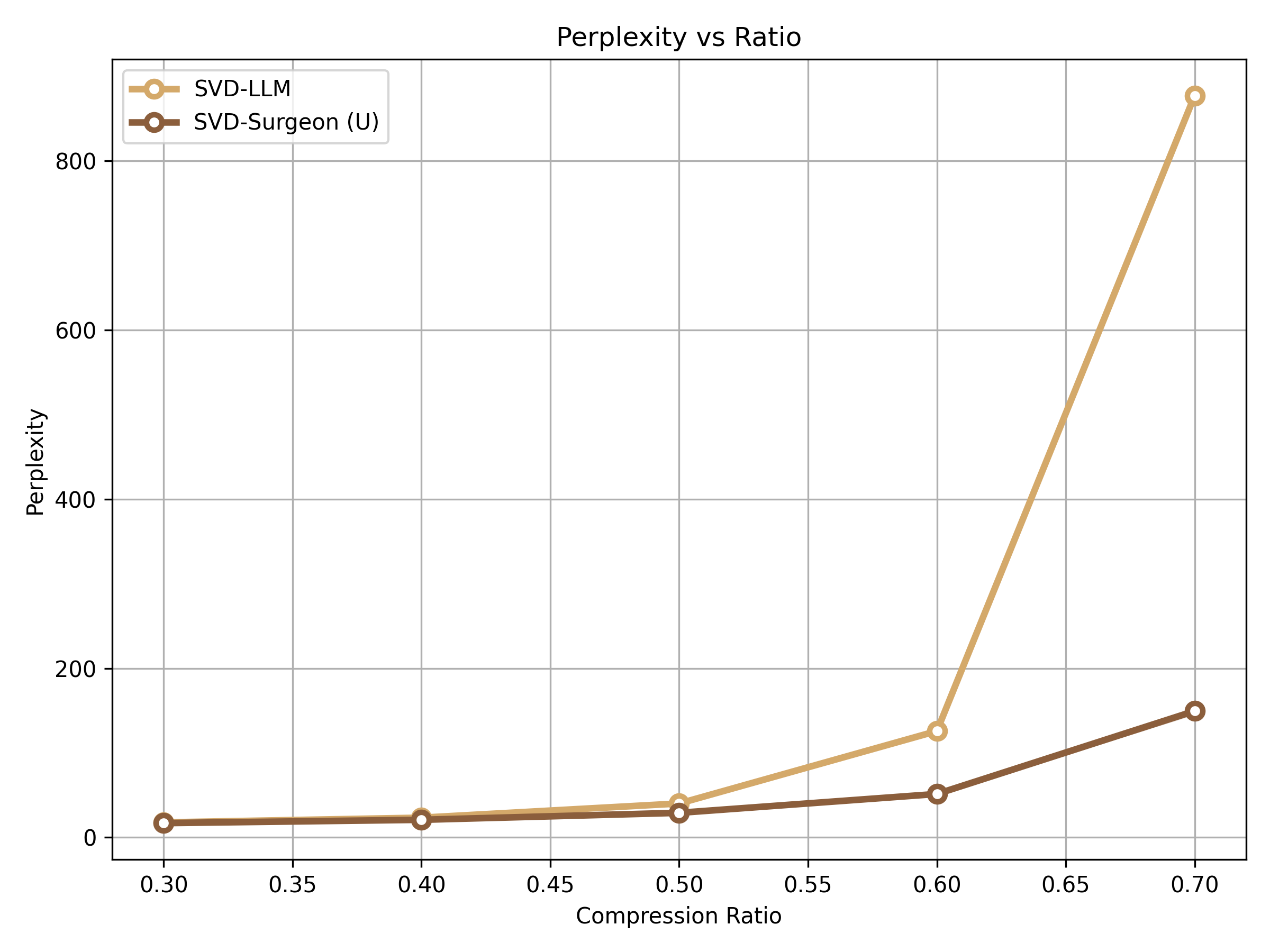}
    \caption{OPT-2.7B}
    \label{fig:ppl_curve_opt_2.7b}
\end{subfigure}

\vspace{0.4cm}

\begin{subfigure}[b]{0.48\textwidth}
    \centering
    \includegraphics[width=\textwidth]{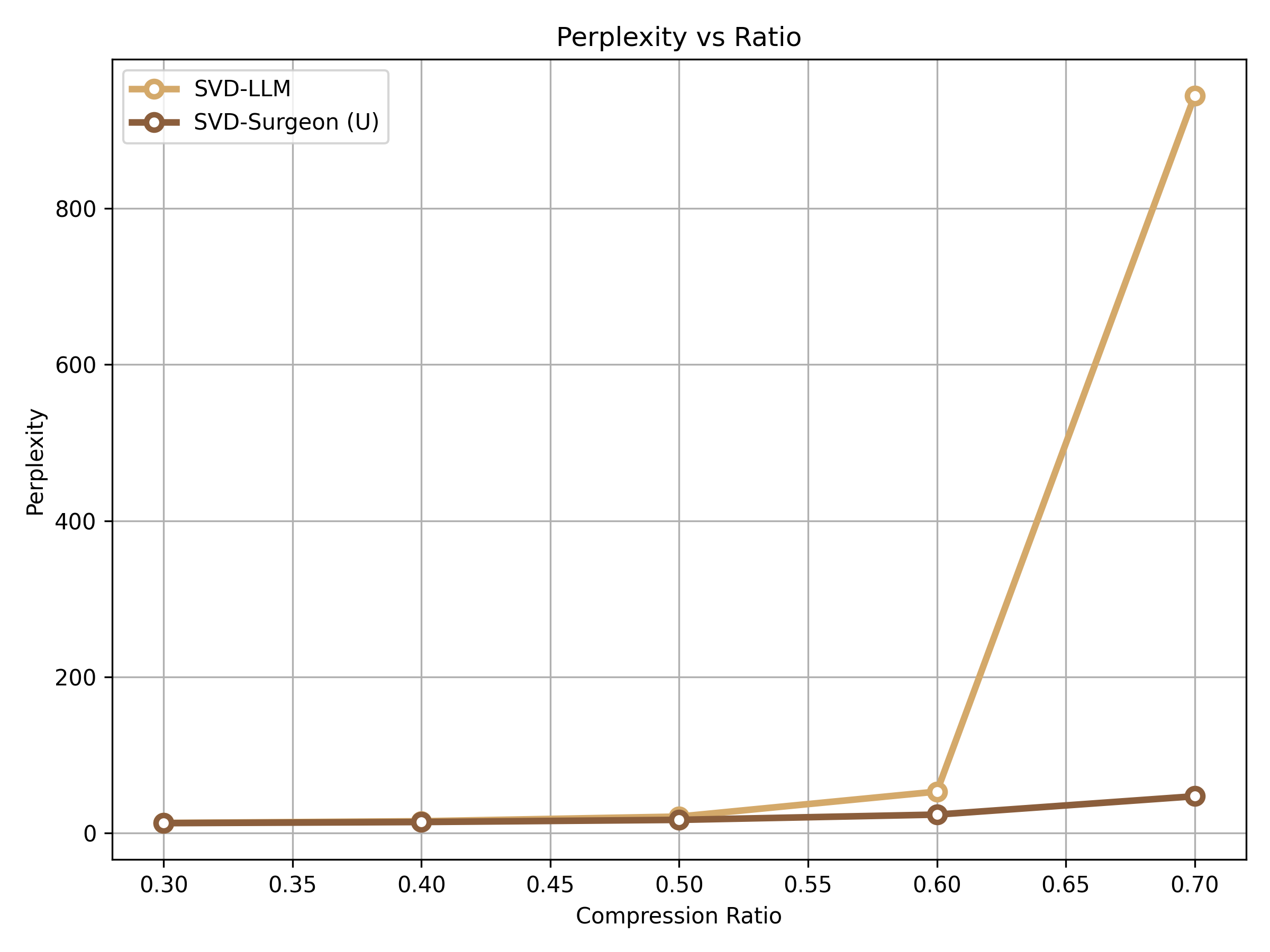}
    \caption{OPT-6.7B}
    \label{fig:ppl_curve_opt_6.7b}
\end{subfigure}
\hfill
\begin{subfigure}[b]{0.48\textwidth}
    \centering
    \includegraphics[width=\textwidth]{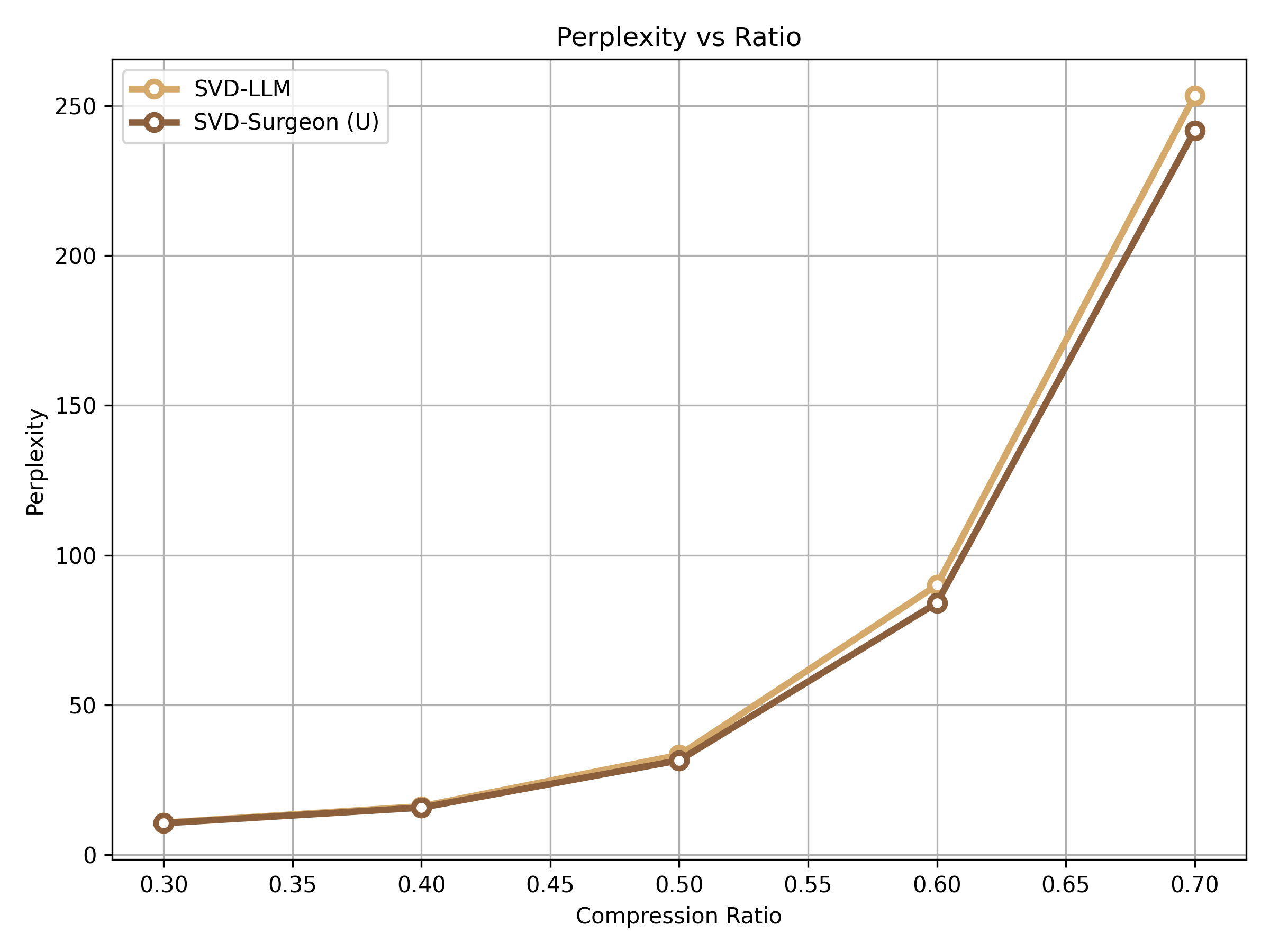}
    \caption{LLaMA-2-7B}
    \label{fig:ppl_curve_llama_2_7b}
\end{subfigure}
\caption{WikiText-2 perplexity vs.\ compression ratio for all four models.
SVD-Surgeon~(U) is shown, while variant~(S) is omitted as it closely tracks~(U)
at this scale. See Table~\ref{tab:ppl_vs_ratio_wiki} for per-ratio
values.} 
\label{fig:ppl_curves}
\end{figure}

Figure~\ref{fig:ppl_curves} plots WikiText-2 perplexity as a function of
compression ratio for all four models. We restrict the range to
$0.3$--$0.7$ for visual clarity. The improvements are most apparent at
aggressive compression, where the curves separate. 
Only the
update-only variant~(U) is shown. The variant~(S) tracks it too closely to be
distinguishable at this scale. Precise per-ratio values are given in
Table~\ref{tab:ppl_vs_ratio_wiki}. Table~\ref{tab:ppl_vs_ratio_wiki_std} additionally reports standard
deviations over three seeds for the OPT family. 

\begin{table}[h]
{\tiny
\setlength{\tabcolsep}{3pt}
\centering
\caption{WikiText-2 perplexity ($\downarrow$) vs.\ compression ratio
(mean $\pm$ std over 3 seeds).
\textbf{Bold}: best; \underline{underline}: second best.}
\label{tab:ppl_vs_ratio_wiki_std}
\begin{tabular}{llccccccc}
\toprule
Model & Method & 0.2 & 0.3 & 0.4 & 0.5 & 0.6 & 0.7 & 0.8 \\
\midrule
\multirow{3}{*}{OPT-1.3B}
  & SVD-LLM          & 17.82 & 20.69 & 27.28 & 47.79 & 140.82 & 654.89 & 4206.70 \\
  & SVD-Surgeon (U)   & $\underline{17.50}{\scriptstyle\pm0.01}$ & $\underline{19.62}{\scriptstyle\pm0.01}$ & $\underline{23.37}{\scriptstyle\pm0.03}$ & $\mathbf{31.29}{\scriptstyle\pm0.05}$ & $\underline{53.63}{\scriptstyle\pm0.04}$ & $\underline{168.48}{\scriptstyle\pm3.43}$ & $\underline{1728.88}{\scriptstyle\pm76.79}$ \\
  & SVD-Surgeon (S)   & $\mathbf{17.49}{\scriptstyle\pm0.01}$ & $\mathbf{19.59}{\scriptstyle\pm0.02}$ & $\mathbf{23.31}{\scriptstyle\pm0.04}$ & $\underline{31.39}{\scriptstyle\pm0.07}$ & $\mathbf{53.46}{\scriptstyle\pm0.17}$ & $\mathbf{163.41}{\scriptstyle\pm1.59}$ & $\mathbf{1587.61}{\scriptstyle\pm111.53}$ \\
\midrule
\multirow{3}{*}{OPT-2.7B}
  & SVD-LLM          & 15.21 & 17.80 & 23.39 & 40.19 & 125.86 & 877.25 & 4584.67 \\
  & SVD-Surgeon (U)   & $\underline{14.93}{\scriptstyle\pm0.02}$ & $\underline{16.97}{\scriptstyle\pm0.03}$ & $\underline{20.91}{\scriptstyle\pm0.01}$ & $\underline{29.15}{\scriptstyle\pm0.09}$ & $\underline{51.43}{\scriptstyle\pm0.37}$ & $\underline{149.52}{\scriptstyle\pm3.08}$ & $\underline{1041.95}{\scriptstyle\pm26.76}$ \\
  & SVD-Surgeon (S)   & $\mathbf{14.83}{\scriptstyle\pm0.02}$ & $\mathbf{16.87}{\scriptstyle\pm0.03}$ & $\mathbf{20.85}{\scriptstyle\pm0.02}$ & $\mathbf{29.11}{\scriptstyle\pm0.09}$ & $\mathbf{51.33}{\scriptstyle\pm0.29}$ & $\mathbf{146.67}{\scriptstyle\pm1.84}$ & $\mathbf{984.74}{\scriptstyle\pm24.76}$ \\
\midrule
\multirow{3}{*}{OPT-6.7B}
  & SVD-LLM          & 12.05 & 13.08 & 15.27 & 21.22 & 53.23 & 944.57 & 6777.57 \\
  & SVD-Surgeon (U)   & $\mathbf{12.00}{\scriptstyle\pm0.01}$ & $\underline{12.81}{\scriptstyle\pm0.01}$ & $\underline{14.25}{\scriptstyle\pm0.01}$ & $\underline{17.02}{\scriptstyle\pm0.01}$ & $\underline{23.76}{\scriptstyle\pm0.08}$ & $\underline{47.27}{\scriptstyle\pm0.10}$ & $\underline{316.47}{\scriptstyle\pm6.31}$ \\
  & SVD-Surgeon (S)   & $\underline{12.01}{\scriptstyle\pm0.01}$ & $\mathbf{12.80}{\scriptstyle\pm0.01}$ & $\mathbf{14.22}{\scriptstyle\pm0.01}$ & $\mathbf{16.90}{\scriptstyle\pm0.02}$ & $\mathbf{23.39}{\scriptstyle\pm0.07}$ & $\mathbf{46.36}{\scriptstyle\pm0.05}$ & $\mathbf{279.88}{\scriptstyle\pm8.63}$ \\
\bottomrule
\end{tabular}}
\end{table}

\end{document}